\documentclass{article}

\usepackage[final, nonatbib]{neurips_2020}

\usepackage[utf8]{inputenc} % allow utf-8 input
\usepackage[T1]{fontenc}    % use 8-bit T1 fonts
\usepackage{hyperref}       % hyperlinks
\usepackage{url}            % simple URL typesetting
\usepackage{booktabs}       % professional-quality tables
\usepackage{amsfonts}       % blackboard math symbols
\usepackage{nicefrac}       % compact symbols for 1/2, etc.
\usepackage{microtype}      % microtypography

\usepackage{amsmath}
\usepackage{multirow}
\usepackage[ruled,vlined]{algorithm2e}
\usepackage{graphicx}
\usepackage{subcaption}

\newcommand{\myalphfoot}
{
\renewcommand{\thefootnote}{\alph{footnote}}
}

\title{A Metric for Linear Symmetry-Based Disentanglement}

\author{
  
  Luis A. P\'{e}rez Rey\thanks{These authors contributed equally}\myalphfoot\footnotemark[1] \hspace{0.4mm}\footnotemark[2] , 
  Loek Tonnaer$^*$\footnotemark[1] \\
  \myalphfoot
  \textbf{Vlado Menkovski\hspace{0.5mm}\footnotemark[1]\hspace{1mm},}\hspace{0.4mm}
  \textbf{Mike Holenderski\hspace{0.5mm}\footnotemark[1]\hspace{1mm},}\hspace{0.4mm}
  \textbf{Jacobus W. Portegies\hspace{0.5mm}\footnotemark[1]}
  \\
  \myalphfoot
\footnotemark[1]\hspace{2.0mm}Eindhoven University of Technology, Eindhoven, The Netherlands \\
\myalphfoot\footnotemark[2]\hspace{2.0mm}Prosus, Amsterdam, The Netherlands \\
  \texttt{\{l.a.perez.rey, l.m.a.tonnaer\}@tue.nl}
}

\begin{document}

\maketitle

% \begin{alphafootnotes}
% \footnotetext[1]{ Eindhoven University of Technology, Eindhoven, The Netherlands}
% \footnotetext[2]{ Prosus, Amsterdam, The Netherlands}
% \end{alphafootnotes}

\begin{abstract}
  The definition of Linear Symmetry-Based Disentanglement (LSBD) proposed by Higgins \textit{et al.} \cite{Higgins2018a} outlines the properties that should characterize a disentangled representation that captures the symmetries of data. However, it is not clear how to measure the degree to which a data representation fulfills these properties. We propose a metric for the evaluation of the level of LSBD that a data representation achieves. We provide a practical method to evaluate this metric and use it to evaluate the disentanglement of the data representations obtained for three datasets with underlying $SO(2)$ symmetries. 
\end{abstract}

% Short description:

% \begin{enumerate}
%     \item LSBD defines the disentanglement of data representations with respect to symmetries. 
%     \item There is no metric to evaluate the level of LSBD of a data representation. 
%     \item We propose a practical method for approximating the metric. 
%     \item We test the approximation on a couple of data representation examples. 
% \end{enumerate}

\section{Introduction}

Disentangled representation learning aims to create low-dimensional representations of data that separate the underlying factors of variation. These representations provide an interpretable \cite{Kaur2019} and useful tool for various purposes, such as noise removal \cite{Lopez2018}, continuous learning \cite{Achille2018}, and visual reasoning \cite{VanSteenkiste2019}. However, there is no consensus about the exact properties that characterize a disentangled representation. Higgins \textit{et al.} \cite{Higgins2018a} provide a formal definition for Symmetry-Based Disentanglement (SBD) and Linear SBD (LSBD) for data representations, building upon the idea that representations should reflect the underlying structure of the data. In particular, they argue that variability in the data comes from symmetry transformations in the real world from which the data is obtained. A formal definition of disentanglement can serve as a paradigm for the evaluation of disentangled representations.

Although several methods have been proposed to capture the transformation properties within data \cite{Cohen2015,Worrall2017,Detlefsen2019}, only few attempt to learn SBD or LSBD representations \cite{Quessard2020, Caselles-Dupre2019}. Moreover, none of them provide a metric for quantifying the level of SBD or LSBD in these representations. In this paper, we propose a metric to characterize the level of LSBD that a data representation achieves, provided that a suitable dataset is available for evaluation.

\section{Linear Symmetry-Based Disentanglement}
The definition of Linear Symmetry-Based Disentangled (LSBD) data representations by Higgins \textit{et al.} \cite{Higgins2018a} formalizes which properties disentangled representations should have to capture the symmetries of data in a linear way, see Appendix \ref{app:lsbd_def}. The definition considers that data is generated from a set of world states whose transformations are modeled by a group $G$ that can be decomposed as the direct product of $K$ groups $G = G_1\times \cdots \times G_K$. The data observed from those world states is represented as elements of a dataspace $X$. A data representation is modeled by an encoding function $h: X\rightarrow Z$ that maps observations in the data $X$ to their representation in the latent space space $Z$.

%$W$ through observations determined by a function $b:W\rightarrow X$ which maps world states from $W$ into datapoints within the data space $X$.

We assume that we can identify a group action\footnote{To simplify the notation we will use $g\cdot x$ for the group actions $G_X(g,x)$ where the set $X$ upon which the group acts can be inferred from context.} $G_X: G\times X \rightarrow X$ of group $G$ on the data space $X$. For example, this can be done if the process of generating data from the real world is bijective \cite{Higgins2018a}.
The encoding function $h$ is LSBD with respect to the group decomposition $G = G_1\times \cdots \times G_K$ if  
\begin{enumerate}
    \item there is a group representation $\rho:G\rightarrow GL(Z)$, with $GL(Z)$ the general linear group on $Z$, that is a \emph{linearly disentangled representation}  with respect to $G = G_1\times \cdots \times G_K$ (see the definition of linearly disentangled representations in Appendix \ref{app:lin_dis_rep}), and
    \item the encoding $h$ is \emph{equivariant} with respect to the action of $G$ on both $X$ and $Z$, meaning that for any $g\in G$ and $x\in X$ 
    \begin{equation}
        h(g\cdot x) = \rho(g)h(x).
    \end{equation}
\end{enumerate}
To measure the level of LSBD of an encoding $h$, we need to measure the level of equivariance of $h$ w.r.t. the most suitable linearly disentangled representation on $Z$. The main questions we address are then how to quantify equivariance, and how to define and find the most suitable representation.

%To measure the level of LSBD of an encoding $h$ we need to measure whether the data representations have similar properties to those of data (equivariance) with respect to a certain linearly disentangled representation that decomposes the encoded data from the latent space $Z$ into invariant vector subspaces $Z = Z_1\oplus\ldots\oplus Z_K$ .
% models the transformations in  whether there is a linearly disentangled representation that decomposes the encoded data from the latent space $Z$ into invariant vector subspaces $Z = Z_1\oplus\ldots\oplus Z_K$ such that the data representations have similar transformation properties to those of the data. 

\section{Metric for Linear Symmetry-Based Disentanglement}
We propose a metric to measure the level of Linear Symmetry-Based Disentanglement (LSBD) of any encoding $h:X \rightarrow Z$ given a data probability measure $\mu$ on $X$, provided that $\mu$ can be written as the pushforward $G_X(\cdot, x_0)_\# \nu$ of some probability measure $\nu$ on $G$ by the function $G_X(\cdot, x_0)$ for some base point $x_0$. More formally,
\begin{equation}
\mu(A) = G_X (\cdot, x_0)_\# \nu(A) = \nu\left(\left\{ g\in G \ | \  G_X(g, x_0)\in A\right\}\right),
\end{equation}
for Borel subsets $A \subset X$. Note that this is only possible if the action $G_X$ is transitive.

For example, the situation of a dataset with $N$ datapoints $\{x_n\}_{n=1}^N = \{ g_n\cdot x_0 \}_{n=1}^N$ corresponds to the case in which $\nu$ and $\mu$ are following empirical measures on the group $G$ and data space $X$, respectively:
\begin{equation}\label{eq:empirical_measures}
\nu := \frac{1}{N} \sum_{i=1}^N \delta_{g_i}, \qquad \mu := \frac{1}{N} \sum_{i=1}^N \delta_{x_i}.
\end{equation}

We define the LSBD metric $\mathcal{M}_\mathrm{LSBD}$ for an encoding $h$ and a measure $\mu$ as
\begin{equation}\label{eq:metric}
\mathcal{M}_\mathrm{LSBD} :=\inf_{\rho \in \mathcal{P}(G, Z)}
\int_G \left\| \rho(g)^{-1} \cdot h(g \cdot x_0) - \int_G \rho(g')^{-1}\cdot h(g'\cdot x_0) d\nu(g') \right\|_{\rho, h, \mu}^2 d \nu(g),
\end{equation}
where the norm $\| \cdot \|_{\rho, h, \mu}$ is a Hilbert-space norm depending on the representation $\rho$, the encoding map $h:X\rightarrow Z$, and the data measure $\mu$. More details of this norm can be found in Appendix \ref{app:inner}. Moreover, $\mathcal{P}(G,Z)$ denotes the set of \emph{linearly disentangled representations} of $G$ in $Z$. Lower values of $\mathcal{M}_{LSBD}$ indicate better disentanglement, zero being optimal.

% The integral whose infimum is calculated in Equation \ref{eq:metric} measures on average how much does the data encodings provided by $h$ differ from the equivariance condition for a certain linearly disentangled group action on $Z$. 

The metric can be interpreted as the measurement of the average deviation from equivariance for the data encodings provided by $h$ with respect to the best linearly disentangled group representation that can be fitted on those encodings. Appendix \ref{app:equiv_m_lsbd} shows alternative expression of $\mathcal{M}_{LSBD}$ metric that makes the measurement of the equivariance property of $h$ more apparent.

\section{Computation of the Linear Symmetry-Based Disentanglement Metric}
In this section we present a method to estimate an upper bound to $\mathcal{M}_{LSBD}$ that gives a good approximation to the metric, for a dataset $\mathcal{X}$ that was generated by some known transformations modeled by the group $G = G_1\times \ldots\times G_K$. Consider that for each subgroup $G_k$ with $k \in \{1,\ldots, K\}$ we have a set of unique and uniformly distributed group elements $\mathcal{G}_k$ that describe all the datapoints in $\mathcal{X}$ in terms of a base point $x_0\in X$ as 
\begin{equation}
\mathcal{X} = \{(g_1, \ldots, g_K)\cdot x_0\in X \ | \ \forall g_k\in \mathcal{G}_k, k \in \{1,\ldots, K\} \} .
\end{equation}
Notice that this dataset follows the empirical measure in Equation (\ref{eq:empirical_measures}). Such a dataset and the information about the underlying group transformations from each $\mathcal{G}_k$ can be used to compute an upper bound to the disentanglement metric $\mathcal{M}_{LSBD}$ for an arbitrary encoding function $h: X\rightarrow Z$.

%An important challenge is to identify for the embedded data encoded in $Z$ by $h$ whether there are invariant subspaces in which a disentangled group representation $\rho$ with respect to $G = G_1\times \ldots, G_K$ can be found.

% In order to compute the level of disentanglement of $h:X\rightarrow Z$ using the metric from Equation \ref{eq:metric} with respect to the dataset $\mathcal{X}$ we need to find a way to express all the possible group representations $\rho\in\mathcal{P}(G,Z)$ that are linearly disentangled. This means that we should identify the decomposition of the latent space $Z$ into invariant subspaces with respect to the group decomposition $G = G_1\times \ldots, G_K$.

Similar to \cite{Cohen2014} we propose to find the disentangled group representations $\rho\in \mathcal{P}(G,Z)$ of $G$ by finding a suitable change of basis that exposes the invariant subspaces into which $Z$ can be decomposed. For the new projected encodings, the group representations can be expressed in a simple way such that the integrals from Equation (\ref{eq:metric}) can be easily approximated with the empirical measures.

A short summary for the process used to calculate the approximation of $\mathcal{M}_{LSBD}$ is as follows. First we find a suitable basis to project the encodings in $Z$ obtained from $\mathcal{X}$ through $h$ such that we can express the group representation as parameterized block-rotational matrices. Then, we estimate an upper bound to $\mathcal{M}_{LSBD}$ for a group representation by approximating the integrals in Equation (\ref{eq:metric}) using the empirical measures. Finally, we tighten this bound by finding the optimal parameters within a search space for the parametric group representations. 
% for the latent variables   $\rho = \boldsymbol{W}\boldsymbol{R}(g)\boldsymbol{W}^T\in \mathcal{P}(G,Z)$. Where $\boldsymbol{W}$ is a fixed orthogonal matrix which indicates a change of basis and $\boldsymbol{R}(g)$ is a block-diagonal rotation matrix that depends on the group element $g\in G$.

We illustrate the steps needed to calculate the upper bound for an example dataset described by a group $G =G_1\times G_2$ with $G_k = SO(2)$ for $k\in\{1,2\}$. For this particular example, assume that the latent space is a $D$-dimensional Euclidean space $Z = \mathbb{R}^D$, with $D\geq 4$. We will denote the encoded data from $\mathcal{X}$ as $z_{g_1, g_2}= h((g_1,g_2)\cdot x_0)$, corresponding to  $g_1\in \mathcal{G}_1$, and $g_2\in \mathcal{G}_2$. For these encodings, the steps to estimate the upper bound to $\mathcal{M}_{LSBD}$ are:

\begin{enumerate}
    \item For each subgroup $G_k$ with $k\in\{1,2\}$ construct a set $\mathcal{Z}_k$ of centered latent representations in which most variability in the embeddings should be attributed to only the $k$-th group acting on the latent representations. For a given $g_1\in \mathcal{G}_1$ and $g_2\in \mathcal{G}_2$ we define
\begin{equation}
  z^{(1)}_{g_1,g_2} :=  z_{g_1,g_2}- \frac{1}{|\mathcal{G}_{2}|}\sum_{g\in \mathcal{G}_{2}} z_{g_1,g} \qquad  z^{(2)}_{g_1,g_2} := z_{g_1,g_2}- \frac{1}{|\mathcal{G}_{1}|}\sum_{g\in \mathcal{G}_{1}} z_{g,g_2}
\end{equation}
such that each $\mathcal{Z}_{k}$ is given by\begin{equation}
    \mathcal{Z}_k = \left\{z^{(k)}_{g_1,g_2} \in Z  \ \middle| \,\forall g_1\in \mathcal{G}_1; \forall g_2\in \mathcal{G}_2 \right\}.
\end{equation}
\item  In our example, the group representations for each $G_k = SO(2)$ are at most 2-dimensional. We use Principal Component Analysis (PCA) on the standardized embeddings of each set $\mathcal{Z}_k$ to find the matrix $\boldsymbol{W}_k$, which corresponds to the orthogonal projection of the encodings from $\mathcal{Z}_k$ into the space spanned by its two first principal components. We use the projected encodings to create a set $\mathcal{Z}'_k$ of 2-dimensional vectors 
\begin{equation}
    \mathcal{Z}'_{k} = \left\{z'^{(k)}_{g_1,g_2} = \frac{1}{\sqrt{2}}\boldsymbol{W}_k \left(z^{(k)}_{g_1,g_2}\odot \frac{1}{\sigma^{(k)}}\right)\in \mathbb{R}^2 \ \middle| \ \forall g_1\in \mathcal{G}_1, \forall g_2\in \mathcal{G}_2 \right\},
\end{equation}
with $\sigma^{(k)}$ the standard deviation vector for the set $\mathcal{Z}_k$. A new set $\mathcal{Z}'$ of 4-dimensional embeddings  is constructed where each element $z'_{g_1, g_2} = (z'^{(1)}_{g_1,g_2},z'^{(2)}_{g_1,g_2})\in \mathcal{Z}'\subseteq \mathbb{R}^4$ for $g_1\in \mathcal{G}_1$ and $g_2\in \mathcal{G}_2$. For this set we will be able to express the group representations as block-diagonal rotation matrices.

\item For the new projected encodings in $\mathcal{Z}'$, the group representation $\rho_{\omega_1, \omega_2}$  of $g =  (g_1, g_2)\in G$ is described as a $4\times 4$ block-diagonal matrix. This matrix is parameterized by a pair of integer values $\omega_1, \omega_2 \in \mathbb{Z}$ as
\begin{equation}
    \rho_{\omega_1,\omega_2}((g_1, g_2)) =    
    \begin{pmatrix}
     \boldsymbol{R}(\omega_1\theta_{g_1})& \\
     & \boldsymbol{R}(\omega_2\theta_{g_2}).
    \end{pmatrix}.
\end{equation}
Each rotation $\boldsymbol{R}$ is a $2\times 2$ matrix that depends on the angle $\theta_{g_{k}}$, which can be identified with each element $g_k$ in the corresponding set $\mathcal{G}_k$. %These parameterized group representations are used to calculate an upper bound to the LSBD metric. 

% \begin{equation}
%     \boldsymbol{R}(\theta)=    
%     \begin{pmatrix}
%     \cos{(\theta)}& -\sin{(\theta)}\\
%     \sin{(\theta)} & \cos{(\theta)}.
%     \end{pmatrix}
% \end{equation}

\begin{figure}[t!]
\centering
\begin{minipage}{0.29\columnwidth}
  \centering
\includegraphics[width=\textwidth]{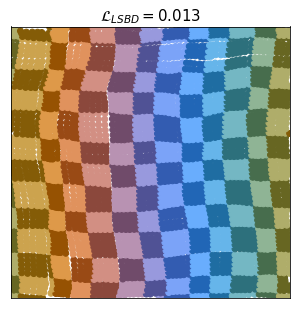}
% \subcaption{$\mathcal{L}_{LSBD} = 0.037$}
\end{minipage}
\centering
\begin{minipage}{0.29\columnwidth}
  \centering
\includegraphics[width=\textwidth]{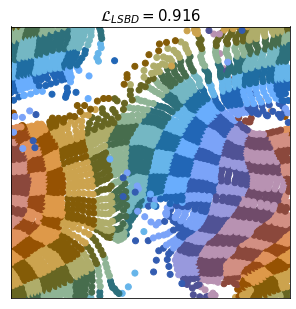}
% \subcaption{$\mathcal{L}_{LSBD} = 0.916$}
\end{minipage}
\centering
\begin{minipage}{0.29\columnwidth}
  \centering
\includegraphics[width=\textwidth]{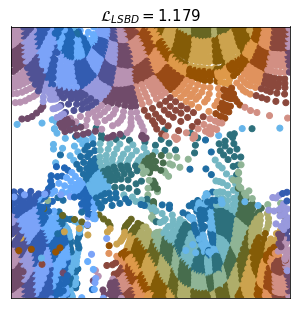}
% \subcaption{$\mathcal{L}_{LSBD} = 1.179 $}
\end{minipage}
  \caption{Plots of the encodings obtained for the Square Translation dataset for three different runs of the Diffusion VAE with $Z = S^1\times S^1\subseteq \mathbb{R}^4$. The plots show the unfolded torus and each axis indicates the angle associated to each circle. The color pattern indicates the correct order that embeddings should have with respect to the underlying group transformations, the borders are periodic. Notice that the least distorted embeddings receive the lowest values for $\mathcal{L}_{LSBD}$ as expected.}
  \label{fig:levels_of_disentanglement}
\end{figure}

\item Finally, we need to approximate the integrals in Equation (\ref{eq:metric}). Recall that the empirical measure for the available group transformations that describe our dataset is expressed as
\begin{equation}
    \nu := \frac{1}{|\mathcal{G}_1|\cdot|\mathcal{G}_2|}\sum_{g_1\in\mathcal{G}_1}\sum_{g_2\in\mathcal{G}_2}\delta_{g_1\times g_2}.
\end{equation}
The integral shown within the $\mathcal{M}_{LSBD}$ in Equation (\ref{eq:metric}) can be approximated by the sum
\begin{equation}
\begin{aligned}
     \overline{z} = \frac{1}{|\mathcal{G}_1|\cdot|\mathcal{G}_2|}\sum_{g_1\in \mathcal{G}_1, }\sum_{g_2\in \mathcal{G}_2}\rho_{\omega_1,\omega_2}^{-1}((g_{1}, g_2))\cdot z'_{g_1, g_2} \approx\int_G \rho(g')^{-1}\cdot h(g'\cdot x_0) d\nu(g'), 
\end{aligned}
\end{equation}
For a fixed pair of integers $\omega_1, \omega_2 \in \mathbb{Z}$ we can approximate an upper bound $\mathcal{L}_{LSBD}(\omega_1, \omega_2)$ to $\mathcal{M}_{LSBD}$ as
\begin{equation}
\begin{aligned}
\mathcal{L}_{LSBD}(\omega_1, \omega_2)
& = 
\frac{1}{|\mathcal{G}_1|\cdot|\mathcal{G}_2|}\sum_{g_1\in \mathcal{G}_1, }\sum_{g_2\in \mathcal{G}_2}\|\rho_{\omega_1,\omega_2}^{-1}((g_1,g_2))\cdot z'_{g_1,g_2}- \overline{z}\|^2 \\
&\approx\int_G \left\| \rho(g)^{-1} \cdot h(g \cdot x_0) - \overline{z} \right\|_{\rho, h, \mu}^2 d \nu(g).
\end{aligned}
\end{equation}
Let $\Omega\subset \mathbb{Z}^2$ be a set of integer pairs. We can tighten the upper bound by finding the integer pair $(\omega_1, \omega_2)\in \Omega$ that minimizes $\mathcal{L}_{LSBD}$.
\begin{equation}\label{eq:upper_bound}
    \mathcal{M}_{LSBD}\leq \min_{(\omega_1, \omega_2)\in \Omega}\mathcal{L}_{LSBD}(\omega_1, \omega_2) = \mathcal{L}_{LSBD}
\end{equation}
\end{enumerate}
\section{Evaluation on Data}\label{sec:evaluation_data}
We estimate the value of $\mathcal{L}_{LSBD}$ for the embeddings of three datasets (Appendix \ref{app:datasets}) with group decomposition $G = SO(2)\times SO(2)$. The  embeddings are obtained by training a standard Variational Autoencoder (VAE) \cite{Kingma2014} with Euclidean latent space $Z = \mathbb{R}^4$ and a Diffusion VAE \cite{PerezRey2020} with $Z = S^1\times S^1\subseteq \mathbb{R}^4$. The quantitative results are shown in Appendix \ref{app:qual_quant}. Figure \ref{fig:levels_of_disentanglement} shows some qualitative and quantitative examples of different degrees of LSBD achieved by the Diffusion VAE. 

As a sanity check, we also evaluated the disentanglement of directly embedding the group elements $g_1\in\mathcal{G}_1$ and $g_2\in\mathcal{G}_2$ in $Z = S^1\times S^1$  with $z_{g_1,g_2} = (\cos{\theta_{g_1}}, \sin{\theta_{g_1}}, \cos{\theta_{g_2}}, \sin{\theta_{g_2}})$. The value $\mathcal{L}_{LSBD}$ for this "perfect" embeddings was of the order of the machine's epsilon (virtually zero) as expected, even after applying a random linear transformation to the perfect embeddings.

\section{Conclusion \& Future Work}
In this paper we introduce a metric to estimate the level of LSBD that a data representation achieves and propose a practical method to approximate the metric, which can be evaluated using a dataset that contains information about its underlying symmetries. In particular, we have evaluated the LSBD of several data representations obtained for datasets with underlying $SO(2)$ symmetries.  

As future work we propose to test and extend the approximations of our metric to more Lie groups such that we can provide more flexibility to the evaluation of LSBD.

\begin{ack}
This work has received funding from the Electronic Component Systems for European Leadership Joint Undertaking under grant agreement No 737459 (project Productive4.0). This Joint Undertaking receives support from the European Union Horizon 2020 research and innovation program and Germany, Austria, France, Czech Republic, Netherlands, Belgium, Spain, Greece, Sweden, Italy, Ireland, Poland, Hungary, Portugal, Denmark, Finland, Luxembourg, Norway, Turkey.

This work has also received funding from the NWO-TTW Programme “Efficient Deep Learning” (EDL) P16-25.

\end{ack}

% \section*{References}

\bibliography{references}
\bibliographystyle{plain}

\appendix

\section{Definition of Linear Symmetry-Based Disentanglement (LSBD) }
\label{app:lsbd_def}
The definition of Higgins \textit{et al.} \cite{Higgins2018a} for LSBD considers data as being generated from a set of world states $W$ through an observation function $b:W\rightarrow X$. The transformations that the world undergoes are modeled by a group $G$ that is the direct product of $K$ groups $G = G_1\times \cdots \times G_K$. A model's internal representation of the data is denoted as the inference map $h:X\rightarrow Z$. The combination of both observation and inference produces a model's internal representation $f: W\rightarrow Z$ of the world, given by the composition $f = h\circ b$.

A model's internal representation $f:W \rightarrow Z$, where $Z$ is a vector space, is LSBD with respect to the group decomposition $G = G_1\times\ldots\times G_K$ if 
\begin{itemize}
    \item there is a decomposition of the representation space $Z = Z_1\oplus \ldots\oplus Z_K$ into $K$ vector subspaces,
    \item there are group representations for each subgroup in the corresponding vector subspace $\rho_k: G_k\rightarrow GL(Z_k)$, $k\in\{1,\ldots,K\}$
    \item the group representation $\rho:G \rightarrow GL(Z)$ for $G$ in $Z$ is linearly disentangled (Appendix \ref{app:lin_dis_rep}), i.e. it acts on $Z$ as
    \begin{equation}
        \rho(g)\cdot z = (\rho_1(g_1)\cdot z_1,\ldots, \rho_K(g_K)\cdot z_K),
    \end{equation}
     for $g = (g_1, \ldots, g_K)\in G$ and $z= (z_1,\ldots, z_K)\in Z$ with $g_k\in G_k$ and $z_k\in Z_k$.
    \item the map $f$ is \emph{equivariant} with respect to the actions of $G$ on $W$ and $Z$, i.e. for all $w\in W$ and $g\in G$ it holds that $f(g\cdot w) = \rho(g)\cdot f(w)$.
\end{itemize}
In this paper we consider the situation where we can construct a group action of $G$ in $X$. This can be achieved easily if the function $b$ is bijective such that for any $g\in G$ and $x\in X$ then $g\cdot x = b(g\cdot b^{-1}(x))$. 

\section{Linearly Disentangled Representations}
\label{app:lin_dis_rep}
We say that a group representation $\rho: G\rightarrow GL(Z)$, with $GL(Z)$ the general linear group on $Z$, is linearly disentangled with respect
to the group decomposition $G = G_1\times \cdots \times G_K$ if there exists a decomposition $Z = Z_1 \oplus\cdots \oplus Z_K$
and representations $\rho_k : G_k \rightarrow GL(Z_k)$ of each group $G_k$ on the corresponding subspace $Z_k$; $k \in\{1,\ldots, K\}$ such that the representation is the direct sum $\rho = \rho_1 \oplus \rho_2 \oplus \cdots\oplus\rho_K$. This means that for a given $g = (g_1,  \cdots, g_K)\in G$ and $z = (z_1,\ldots, z_K)\in Z$ with $g_k\in G_k$ and $z_k\in Z_k$ the group representation of $\rho(g)$ acts on $Z$ according to 
\begin{equation}
\rho((g_1,  \cdots, g_K))(z_1,\ldots, z_K) = (\rho_1(g_1)\cdot v_1, \ldots, \rho_2(g_2)\cdot v_K).    
\end{equation}

\section{Inner Product}\label{app:inner}

To describe the norm $\| \cdot \|_{\rho, h, \mu}$ we start with an arbitrary inner product $(\cdot, \cdot)$ on the linear latent space $Z$.
Assume that $\rho$ is linearly disentangled and accodingly splits in irreducible representations $\rho_k : G \to Z_k$ where $Z = Z_1 \oplus \cdots \oplus Z_K$ for some $K \in \mathbb{N}$.
We will define a new inner product $\langle \cdot, \cdot \rangle_{\rho, h, \mu}$ on $Z$ as follows.
First of all we declare $Z_k$ and $Z_m$ to be orthogonal with respect to $\langle \cdot, \cdot \rangle_{\rho, h, \mu}$ if $k \neq m$. We denote by $\pi_k$ the orthogonal projection on $Z_k$.

For $z, z' \in Z_i$, we set
\begin{equation}
\langle z, z' \rangle_{\rho, h, \mu} := \lambda_{k, h, \mu}^{-1} \int_{g \in G} (\rho(g)\cdot z, \rho(g)\cdot z' ) d \mathfrak{m}(g)
\end{equation}
where $\mathfrak{m}$ is the (bi-invariant) Haar measure normalized such that $\mathfrak{m}(G) = 1$ and set
\begin{equation}
\lambda_{k, h, \mu} := \int_X \int_G \|\pi_k(h(x))\|^2 d \mathfrak{m}(g) d\mu(x)
\end{equation}
if the integral on the right-hand side is strictly positive and otherwise we set $\lambda_k := 1$. This construction completely specifies the new inner product, and it has the following properties:
\begin{itemize}
    \item the subspaces $Z_k$ are mutually orthogonal,
    \item $\rho_k(g)$ is orthogonal on $Z_k$ for every $g \in G$, in other words $\rho_k$ maps to the orthogonal group on $Z_k$. Moreover, $\rho$ maps to the orthogonal group on $Z$. This follows directly from the bi-invariance of the Haar measure and the definition of $\langle\cdot, \cdot\rangle_{\rho, h, \mu}$.
    \item If $\pi_k$ is the orthogonal projection to $Z_k$, then 
    \begin{equation}
    \int_X \| \pi_k(h(x)) \|_{\rho, h, \mu}^2  d\mu(x) = 1
    \end{equation}
    if the integral on the left is strictly positive.
\end{itemize}

For an arbitrary pair $z, z'\in Z$ the inner product $\langle\cdot,\cdot \rangle_{\rho, h, \mu}$ is given by 
\begin{equation}
    \langle z, z'\rangle_{\rho, h, \mu} = \sum_{k = 1}^K \lambda_{k, h, \mu}^{-1}\int_{g\in G} (\rho(g)\cdot \pi_{k}(z), \rho(g)\cdot\pi_{k}(z'))d\mathfrak{m}(g)
\end{equation}

\section{Evaluation of Equivariance by $\mathcal{M}_{LSBD}$}\label{app:equiv_m_lsbd}
We will now give an alternative expression for the disentanglement metric $\mathcal{M}_{LSBD}$, since it will more visibly relate to the definition of equivariance. To avoid notational cluttering, in this section we will denote the norm $\|\cdot\|_{\rho, h, \mu}$ as $\| \cdot \|_*$.
Let $\rho\in \mathcal{P}(G,Z)$ be a linear disentangled representation of $G$ in $Z$. By expanding the inner product (or by using usual computation rules for expectations and variances), we first find that
\begin{equation}
\begin{split}
&\int_G \left\|\rho(g)^{-1}\cdot h(g\cdot x_0) - \int_G \rho(g')^{-1} \cdot h(g'\cdot x_0)d\nu(g') \right\|_*^2 d \nu(g)\\
&= \int_G \left\| \rho(g)^{-1}\cdot h(g\cdot x_0) \right\|_*^2 d\nu(g) - \left\|\int_G \rho(g)^{-1}\cdot h(g\cdot x_0) d\nu(g)\right\|_*^2\\
&= \frac{1}{2}\int_G \int_G \| \rho(g)^{-1}\cdot h(g\cdot x_0) - \rho(g')^{-1} \cdot h(g'\cdot x_0) \|_*^2 d \nu(g) d\nu(g').
\end{split}
\end{equation}
We now use that $\rho$ maps to the orthogonal group for $(\cdot, \cdot)_*$, so that we can write the same expression as
\begin{equation}
\frac{1}{2}\int_G \int_G \| \rho(g\circ g'^{-1})^{-1} \cdot h(((g\circ g'^{-1})\cdot g')\cdot x_0) - h(g'\cdot x_0) \|_*^2 d \nu(g) d\nu(g').
\end{equation}
This brings us to the alternative characterization of $\mathcal{M}_{LSBD}$ as
\begin{equation}
\mathcal{M}_{LSBD} = \inf_{\rho \in \mathcal{P}(G,Z)} \frac{1}{2}\int_G \int_G \|\rho(g\circ g'^{-1})^{-1} h(((g\circ  g'^{-1})\cdot g') \cdot x_0) - h(g'\cdot x_0)\|_{*}^2 d \nu(g) d \nu(g').
\end{equation}
In particular, if for every data point $x$ there is a unique group element $g_x$ such that $x = g_x\cdot x_0$, the disentanglement metric $\mathcal{M}_{LSBD}$ can also be written as
\begin{equation}
\inf_{\rho \in \mathcal{P}(G,Z)} \frac{1}{2}\int_G \int_X \|\rho(g\circ  g_x^{-1})^{-1} h((g\circ g_x^{-1})\cdot x) - h(x)\|_{*}^2 d \nu(g) d \mu(x), 
\end{equation}
in which the equivariance condition appears prominently. The condition becomes even more apparent if $\nu$ is in fact the Haar measure itself, in which case the metric equals
\begin{equation}
\inf_{\rho \in \mathcal{P}(G,Z)} \frac{1}{2}\int_{G} \int_X \|\rho(g)^{-1} \circ h(g\cdot x) - h(x)\|_*^2 d \mathfrak{m}(g) d\mu(x).
\end{equation}
% Although this expression would also make for a natural choice of metric in general (when $\nu$ is not necessarily the Haar measure), this choice is not directly practical as one does not usually have access to data points $g\cdot x$ for arbitrary $g$.

\section{Datasets}
\label{app:datasets}
\begin{figure}[t!]
\centering
\begin{minipage}{0.3\columnwidth}
  \centering
\includegraphics[width=\textwidth]{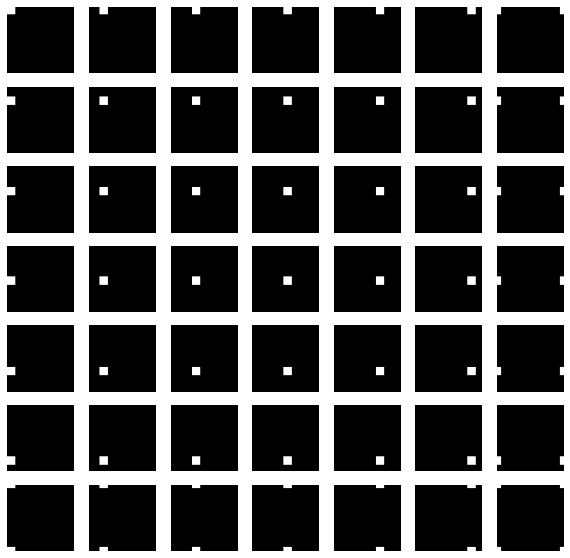}
\subcaption{Square Translation}
\end{minipage}
\centering
\begin{minipage}{0.3\columnwidth}
  \centering
\includegraphics[width=\textwidth]{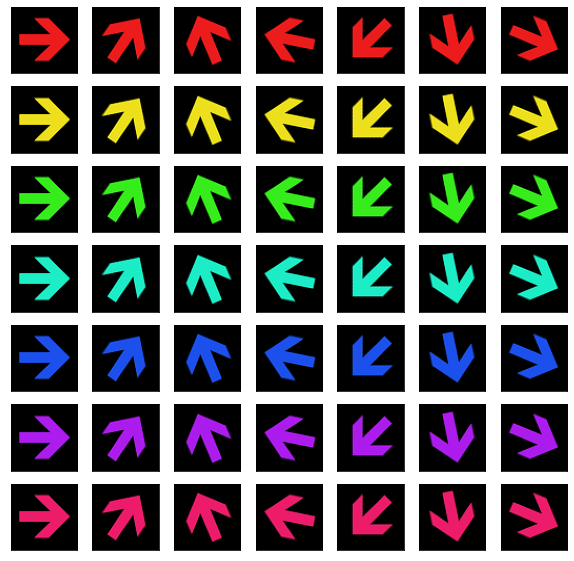}
\subcaption{Arrow Rotation}
\end{minipage}
\centering
\begin{minipage}{0.3\columnwidth}
  \centering
\includegraphics[width=\textwidth]{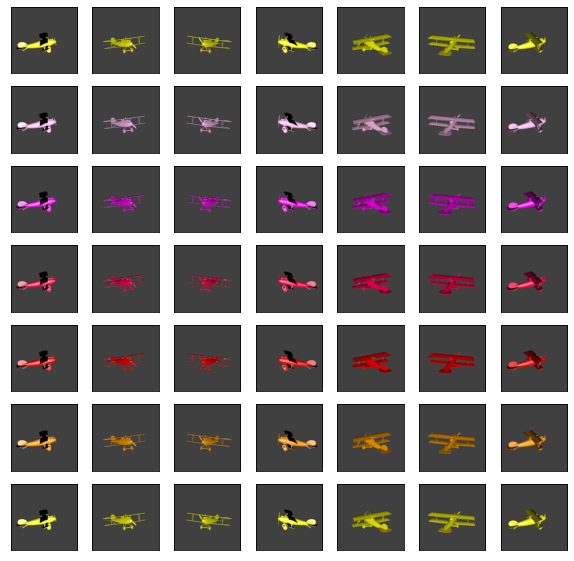}
\subcaption{ModelNet40 Airplane}
\end{minipage}
  \caption{Example images from each of the datasets used. Each image corresponds to an example data point for a combination of two factors, e.g. color and orientation. The factors change horizontally and vertically and the boundaries of each dataset example are periodic.}
  \label{fig:dataset_example}
\end{figure}

All datasets contain $64 \times 64$ pixel images, with a known group decomposition $G = SO(2) \times SO(2)$ describing the underlying transformations. For each subgroup a fixed number of $|\mathcal{G}_k| = 64$ with $k\in \{1,2\}$ transformations is selected. Each image is generated from a single initial data point upon which all possible group actions are applied, resulting in datasets with $|\mathcal{G}_1|\cdot |\mathcal{G}_2| = 4096$ images. The datasets exemplify different group actions of $SO(2)$: periodic translations, in-plane rotations, out-of-plane rotations, and periodic hue-shifts, see Figure \ref{fig:dataset_example}.

\paragraph{Square Translation}
This dataset consists of a set of images of a black background with a square of $8\times8$ white pixels. The dataset is generated applying vertical and horizontal translations of the white square considering periodic boundaries.

\paragraph{Arrow Rotation}
This dataset consists of a set of images depicting a colored arrow at a given orientation. The dataset is generated by applying cyclic shifts of its color and in-plane rotations. The cyclic color shifts were obtained by preselecting a fixed set of $64$ colors from a circular hue axis. The in-plane rotations were obtained by rotating the arrow along an axis perpendicular to the picture plane over $64$ predefined positions.

\paragraph{ModelNet40 Airplane}
The ModelNet40 Airplane dataset consists of a dataset of renders obtained using Blender v2.7 \cite{Community2020} from a 3D model of an airplane within the ModelNet40 dataset \cite{Wu2014}. We created each image by varying two properties: the airplane's color and its orientation with respect to the camera. The orientation was changed via rotation with respect to a vertical axis (out-of-plane rotation). The colors of the model were selected from a predefined cyclic set of colors similar to the arrow rotation dataset.

\section{Quantitative Results}\label{app:qual_quant}
In this section we present the quantitative results in Table \ref{tab:results} from training a standard Variational Autoencoder (VAE) \cite{Kingma2014} with Euclidean latent space $Z = \mathbb{R}^4$ and a Diffusion VAE \cite{PerezRey2020} with embeddings on a flat torus $Z = S^1\times S^1\subseteq \mathbb{R}^4$ embedded in $\mathbb{R}^4$. We estimated $\mathcal{L}_{LSBD}$ by finding the integer pairs from all the combinations in $\Omega = \{-10, \ldots, 10\}^2$
. Each model was trained for 300 epochs and the results for the $\mathcal{L}_{LSBD}$ were averaged across 10 repetitions.

\begin{table}[h]
    \centering
    \begin{tabular}{|c|c|c|c|}
    \hline
    \multirow{2}{*}{Manifold}&\multicolumn{3}{|c|}{$\mathcal{L}_{LSBD}$} 
    
    \\
    \cline{2-4} 
    & Square & Arrow & Airplane\\
    \hline
         $\mathbb{R}^4$ &  $1.958\pm 0.045$&$1.354\pm 0.141$&$1.312\pm0.180$\\
         $S^1\times S^1$&$0.399\pm 0.384$&$1.570\pm 0.152$&$1.834\pm0.100$\\
    \hline
    \end{tabular}
    \caption{Estimated $\mathcal{L}_{LSBD}$ for the embeddings obtained with a standard VAE on an Euclidean latent space $\mathbb{R}^4$ and a Diffusion VAE with hypertoroidal latent space $S^1\times S^1$ across three datasets with underlying group structure $G = SO(2)\times SO(2)$ for 10 repetitions.}
   \label{tab:results}
\end{table}
\end{document}